# Variational Bayesian Inference with Stochastic Search

**John Paisley**[1]  JPAISLEY@BERKELEY.EDU
**David M. Blei**[3]  BLEI@CS.PRINCETON.EDU
**Michael I. Jordan**[1,2]  JORDAN@EECS.BERKELEY.EDU
[1]Department of EECS, [2]Department of Statistics, UC Berkeley
[3]Department of Computer Science, Princeton University

## Abstract

Mean-field variational inference is a method for approximate Bayesian posterior inference. It approximates a full posterior distribution with a factorized set of distributions by maximizing a lower bound on the marginal likelihood. This requires the ability to integrate a sum of terms in the log joint likelihood using this factorized distribution. Often not all integrals are in closed form, which is typically handled by using a lower bound. We present an alternative algorithm based on stochastic optimization that allows for direct optimization of the variational lower bound. This method uses *control variates* to reduce the variance of the stochastic search gradient, in which existing lower bounds can play an important role. We demonstrate the approach on two non-conjugate models: logistic regression and an approximation to the HDP.

## 1. Introduction

Mean-field variational Bayesian (MFVB) inference is an optimization-based approach to approximating the full posterior of the latent variables of a Bayesian model (Jordan et al., 1999). It has been applied to many problem domains, for example mixture modeling (Blei & Jordan, 2006), sequential modeling (Beal, 2003) and factor analysis (Paisley & Carin, 2009). In addition, recent development of the theory has extended the method to online inference and stochastic optimization settings, making variational Bayes a viable approach for Bayesian learning with massive data sets (Hoffman et al., 2010; Wang et al., 2011).

Variational Bayes approximates the full posterior by attempting to minimize the Kullback-Leibler divergence between the true posterior and a predefined factorized distribution on the same variables. Minimizing this divergence is equivalent to maximizing the familiar variational objective function. To review, let $\Theta = \{\theta_i\}$ represent the set of latent variables (random effects and parameters) in the model and $X$ represent the data. The joint likelihood of $X$ and $\Theta$ is $P(X, \Theta|\Upsilon)$, with $\Upsilon$ the set of hyperparameters. Variational inference approximates the posterior $P(\Theta|X, \Upsilon)$ with a $Q$ distribution that takes a set of variational parameters $\Psi = \{\psi_i\}$. This distribution is factorized, $Q(\Theta|\Psi) = \prod_i q_i(\theta_i|\psi_i)$, and the values of $\Psi$ are optimized to maximize the objective function,

$$\mathcal{L}(X, \Psi) = \mathbb{E}_Q[\ln P(X, \Theta|\Upsilon)] + \mathbb{H}[Q(\Theta|\Psi)]. \qquad (1)$$

The solution is only locally optimal when $\mathcal{L}$ is not convex, which is usually the case. Most variational inference algorithms optimize $\mathcal{L}$ by coordinate ascent, which repeatedly cycles through and optimizes with respect to each variational parameter $\psi_i$. Often the locally optimal value of $\psi_i$ has a closed-form solution, for example in conjugate exponential models.

The log of the joint likelihood results in a sum of terms; a major issue that often arises in MFVB is that not all expectations in this sum are in closed form. A typical solution in this case is to replace the problematic function with another function of the same variables (plus auxiliary variables) that is a point-wise lower bound. This new function is selected such that the expectation is tractable. While inference can now proceed, a drawback of introducing bounds is that the true variational objective function is no longer being optimized, which may lead to a significantly worse posterior approximation. Therefore, much attention has been paid to developing tight bounds of commonly occurring functions (e.g., Jaakkola & Jordan (2000), Marlin et al. (2011), Leisink & Kappen (2001)).

We present a method for directly optimizing Eq. (1) for models in which not all expectations are tractable; we show how a stochastic approximation of $\nabla_{\psi_i}\mathcal{L}$ can allow for optimization of $\mathcal{L}$ when the expectation $\mathbb{E}_{q_i}[\ln P(X,\Theta|\Upsilon)]$ is not in closed form. The approximation is unbiased, and so by using the proposed stochastic method we are directly optimizing $\mathcal{L}$.

Our stochastic approximation is based on Monte Carlo integration, for which the number of samples heavily depends on the variance of this approximation. We introduce a *control variate* (Ross, 2006) to significantly reduce the variance of this stochastic approximation. A control variate is a tractable function $g$ that is highly correlated with the intractable function $f$. The function $g$ replaces $f$ in Eq. (1), and the gradient is then stochastically corrected for bias.

Existing lower bounds have properties that make them ideal as control variates, and thus can improve the speed of the algorithm. However, a major advantage of the control variate methodology is that it does not require the tractable function $g$ to bound $f$, but only to correlate well with it (i.e., to approximate it well modulo a scaling). This opens the door to many more functions that may give better approximations than a lower bound. One of these possible functions is the second-order Taylor expansion, which often gives a very good approximation, while also allowing for closed-form expectations. We show the potential performance gain using this function as a control variate, which we denote the *control variate delta method* for MFVB.

**Related work.** Recent work by Knowles & Minka (2011) has also addressed the problem of intractable expectations in MFVB inference in the context of developing a more general variational message passing algorithm. Our solution arises from a different perspective and results in a new algorithm based on stochastic optimization. Graves (2011) considers a similar problem for neural networks, but a lack of control variates limits the algorithm to significantly simpler variational approximations. Stochastic search algorithms have also been developed for models of Evolution Strategies (see, e.g., Yi et al. (2009)).

## 2. Mean-field variational inference

Mean-field variational Bayesian (MFVB) inference approximates the full posterior of the latent variables of a Bayesian model with a factorized distribution. As motivated in the introduction, let $\Theta = \{\theta_i\}$ be these variables, $X$ the data and $\Upsilon$ all hyperparameters of the prior distributions on $\Theta$. We define the factorized distribution on $\Theta$ to be $Q(\Theta|\Psi) = \prod_i q_i(\theta_i|\psi_i)$, where $\psi_i$ are the parameters of the $q_i$ distributions. The variational objective function arises by bounding the marginal likelihood using the $Q$ distribution,

$$\ln P(X|\Upsilon) = \ln \int_\Theta P(X,\Theta|\Upsilon) d\Theta \qquad (2)$$
$$\geq \int_\Theta Q(\Theta|\Psi) \ln \frac{P(X,\Theta|\Upsilon)}{Q(\Theta|\Psi)} d\Theta.$$

Maximizing this lower bound (denoted $\mathcal{L}$) with respect to $\Psi$ is equivalent to minimizing the Kullback-Leibler divergence between $Q(\Theta)$ and $P(\Theta|X,\Upsilon)$, which makes up the difference in Eq. (2).

To facilitate our discussion, we write the functions appearing in the log joint likelihood as $\ln P(X,\Theta|\Upsilon) = \sum_j f_j(X_{A_j},\Theta_{B_j})$, where $A_j$ indexes the data appearing in function $j$ and $B_j$ indexes the latent variables appearing in function $j$. We note that the index $j$ does not correspond to variables or distributions, but to the terms of the log joint likelihood. Using this notation, the variational lower bound in Eq. (1) becomes

$$\mathcal{L} = \sum_j \mathbb{E}_Q[f_j(X_{A_j},\Theta_{B_j})] + \sum_i \mathbb{H}[q_i(\theta_i|\psi_i)]. \quad (3)$$

For each function $f_j$, those $\theta_i \notin \Theta_{B_j}$ will have their corresponding $q_i$ removed from the expectation. For those $\theta_i \in \Theta_{B_j}$, the expectation of $f_j$ results in a new function of variational parameters $\psi_i \in \Psi_{B_j}$. Ideally, all expectations will be in closed form, allowing for the optimization of $\Psi$ to proceed.

In the case where an expectation in Eq. (3) is not tractable, a nicer functional lower bound can replace the problematic function. That is, let $\mathbb{E}_{q_i}[f_j(\theta_i)]$ be intractable.[1] A common approach to dealing with this issue is to introduce a function $g(\theta_i,\xi)$ that replaces $f_j$ and is a point-wise lower bound: $f_j(\theta_i) \geq g(\theta_i,\xi)$ for all $\theta_i$. The function $g$ usually takes auxiliary variables $\xi$, which determines how tightly $g$ approximates $f_j$ and is tuned along with the other parameters during inference. The expectation $\mathbb{E}_{q_i}[g(\theta_i,\xi)]$ has a closed-form solution, and gives a lower bound on the variational objective that can be optimized.

To illustrate, consider the case where $f_j$ is convex in $\theta_i$. Then a bound $g$ could be a first-order Taylor expansion of $f_j$ about the point $\xi$, which has a closed-form expectation. Significantly tighter tractable bounds have also been developed for various frequently occurring functions (e.g., Marlin et al. (2011), Knowles & Minka (2011)). In general, the looser the bound the further one is from optimizing the variational objective, and learning of $\psi_i$ can suffer as a result.

---

[1] We have simplified the notation for clarity.

## 3. Stochastic search variational Bayes

We next present a method based on stochastic search for directly optimizing the variational objective function $\mathcal{L}$ in cases where some expectations cannot be computed in the log joint likelihood. This method uses a stochastic approximation of the gradient with respect to the variational parameters of the associated $q$ distribution. To further simplify notation, we drop all indices; $f$ is the intractable function of $\theta$ (plus other variational parameters), and $\theta$ has a variational distribution $q$ taking parameters $\psi$.

We separate the lower bound $\mathcal{L}$ into two functions, $\mathbb{E}f$ and $h$, where $h(X, \Psi)$ contains everything in $\mathcal{L}$ except for $\mathbb{E}f$. Notably, $h$ contains all other functions of $\psi$ resulting from expectations calculated with respect to $q$. In coordinate ascent variational inference, the first step in optimizing $q$ with respect to its parameters $\psi$ is to take the gradient of the variational objective,

$$\nabla_\psi \mathcal{L} = \nabla_\psi \mathbb{E}_q[f(\theta)] + \nabla_\psi h(X, \Psi). \quad (4)$$

This gradient contains a tractable term resulting from $\nabla_\psi h$, and an intractable term $\nabla_\psi \mathbb{E}_q f$. Our goal is to make a stochastic approximation of this gradient. To this end, assuming the necessary regularity conditions, we rewrite this function as

$$\begin{aligned}
\nabla_\psi \mathbb{E}_q[f(\theta)] &= \nabla_\psi \int_\theta f(\theta) q(\theta|\psi) d\theta \quad (5) \\
&= \int_\theta f(\theta) \nabla_\psi q(\theta|\psi) d\theta \\
&= \int_\theta f(\theta) q(\theta|\psi) \nabla_\psi \ln q(\theta|\psi) d\theta.
\end{aligned}$$

We use the identity $\nabla_\psi q(\theta|\psi) = q(\theta|\psi) \nabla_\psi \ln q(\theta|\psi)$. It follows that $\nabla_\psi \mathbb{E}_q[f(\theta)] = \mathbb{E}_q[f(\theta)\nabla_\psi \ln q(\theta|\psi)]$. We can stochastically approximate this expectation using Monte Carlo integration,

$$\nabla_\psi \mathbb{E}_q[f(\theta)] \approx \frac{1}{S} \sum_{s=1}^{S} f(\theta^{(s)}) \nabla_\psi \ln q(\theta^{(s)}|\psi), \quad (6)$$

where $\theta^{(s)} \stackrel{iid}{\sim} q(\theta|\psi)$ for $s = 1, \ldots, S$. We can therefore replace $\nabla_\psi \mathbb{E}_q[f(\theta)]$ with the unbiased stochastic approximation of this gradient in Eq. (6). Denote this approximation as $\zeta$. At iteration $t$, we update the variational parameter $\psi$ by taking a gradient step,

$$\psi^{(t+1)} = \psi^{(t)} + \rho_t \nabla_\psi h(X, \Psi^{(t)}) + \rho_t \zeta_t. \quad (7)$$

By decreasing the step size $\rho_t$ such that $\sum_{t=1}^\infty \rho_t = \infty$ and $\sum_{t=1}^\infty \rho_t^2 < \infty$, convergence to a local optimal solution of $\mathcal{L}$ is guaranteed. For example, $\rho_t = (w+t)^{-\eta}$ with $\eta \in (0.5, 1]$ and $w \geq 0$ satisfies this requirement.

## 4. Searching with control variates

A practical issue with the stochastic approximation proposed in Sec. 3 is that the variance of the gradient approximation may be very large. Given $S$ samples of a random vector $X$, the covariance of its unbiased sample mean $\bar{X}$ is known to be $\text{Cov}(\bar{X}) = \text{Cov}(X)/S$. When the diagonal values of $\text{Cov}(X)$ are large, many samples will be required to bring this variance below a desired level for approximating the expectation. As our experiments will show in Sec. 6, the value of $S$ can be very large in practice and lead to a slow algorithm. We therefore seek a variance reduction method to reduce the number of samples needed to construct the stochastic search direction.

We introduce a *control variate* (Ross, 2006) to reduce the variance of the stochastic gradient constructed in Eq. (6). A control variate is a random variable that is highly correlated with an intractable variable, but for which the expectation is tractable. In this case the random variable is $f(\theta)$, for which we introduce a control variate $g(\theta)$. Control variates are ideal for MFVB because they can leverage the existing bounds, though they also admit a larger class of functions. We next review this variance reduction technique for $\mathbb{E}f$, and discuss the modifications needed to account for the stochastic vector $f(\theta)\nabla_\psi \ln q(\theta|\psi)$.

### 4.1. A control variate for $f(\theta)$

Generally speaking, variance reduction works by modifying a function of a random variable such that its expectation remains the same, but its variance decreases. Toward this end, we introduce a control variate $g(\theta)$, which approximates $f(\theta)$ well in the highly probable regions as defined by $q(\theta)$, but also has a closed-form expectation under $q$. Using $g$ and a scalar $a \in \mathbb{R}$, we first form the new function $\hat{f}$,

$$\hat{f}(\theta) = f(\theta) - a(g(\theta) - \mathbb{E}_q[g(\theta)]). \quad (8)$$

This function has the same expectation as $f$ and therefore can replace it in $\mathcal{L}$ in Eq. (3).

The next step is to set the value of $a$ to minimize the variance of $\hat{f}$. A simple calculation shows that

$$\text{Var}(\hat{f}) = \text{Var}(f) - 2a\text{Cov}(f, g) + a^2 \text{Var}(g). \quad (9)$$

Taking the derivative with respect to $a$ and setting to zero gives the optimal value,

$$a = \frac{\text{Cov}(f, g)}{\text{Var}(g)}. \quad (10)$$

As is usual, this covariance and variance is unknown in the functions we encounter. We can approximate

---

**Algorithm 1** Variational Bayes with stochastic search

---

**Goal** To calculate $\nabla_\psi \mathcal{L} = \nabla_\psi \mathbb{E}_q[f(\theta)] + \nabla_\psi h(X, \Psi)$.
**Approximate** $\nabla_\psi \mathcal{L}$ using stochastic search.
**input** Variance reduction parameter $\epsilon$.

1: Introduce the function $g(\theta)$ as a control variate that highly correlates with $f(\theta)$.
2: Sample an initial (small) collection $\theta^{(s)} \sim q(\theta|\psi)$.
3: Sum the sample variances and covariances
 $\beta = \sum_{k=1}^{K} \text{Var}(g\frac{\partial \ln q}{\partial \psi_k})$, $\gamma = \sum_{k=1}^{K} \text{Var}(f\frac{\partial \ln q}{\partial \psi_k})$,
 $\alpha = \sum_{k=1}^{K} \text{Cov}(f\frac{\partial \ln q}{\partial \psi_k}, g\frac{\partial \ln q}{\partial \psi_k})$.
4: Set $\hat{a} = \alpha/\beta$ and $S = (\gamma - \alpha^2/\beta)/\epsilon K$.
5: Sample $\theta^{(s)} \sim q(\theta|\psi)$ i.i.d. for $s = 1, \ldots, \lceil S \rceil$.
6: Construct the stochastic search vector
 $\zeta = \frac{1}{S}\sum_{s=1}^{S} \{f(\theta^{(s)}) - \hat{a}g(\theta^{(s)})\}\nabla_\psi \ln q(\theta^{(s)}|\psi)$.
7: Step in the direction of the stochastic gradient
 $\psi' = \psi + \rho\zeta + \rho\nabla_\psi(h(X,\Psi) + \hat{a}\mathbb{E}_q[g(\theta)])$.

---

$a$ with $\hat{a}$, found by plugging the sample variance and covariance into Eq. (10) using samples from the algorithm.

The potential reduction in variance is seen by plugging Eq. (10) into Eq. (9) and taking the ratio of the two variances,

$$\text{Var}(\hat{f})/\text{Var}(f) = 1 - \text{Corr}(f, g)^2. \quad (11)$$

Therefore, the greater the correlation between $f$ and $g$, the greater the variance reduction. Tight lower bounds of $f$ by construction have this high correlation, but we note that tight *upper* bounds work as well, as do well-approximating functions that do not bound $f$.

Using the control variate $g$, we now write the stochastic approximation to the gradient as

$$\nabla_\psi \mathbb{E}_q[\hat{f}(\theta)] \approx \hat{a}\nabla_\psi \mathbb{E}_q[g(\theta)] \quad (12)$$
$$+ \frac{1}{S}\sum_{s=1}^{S}\{f(\theta^{(s)}) - \hat{a}g(\theta^{(s)})\}\nabla_\psi \ln q(\theta^{(s)}|\psi),$$

where $\theta^{(s)} \overset{iid}{\sim} q(\theta|\psi)$ for $s = 1, \ldots, S$.

Writing the stochastic approximation this way allows for a more intuitive understanding of the algorithm. By separating the tractable and stochastic parts as done in Eq. (12), we first replace the intractable function $f$ with a tractable approximation $g$. (This resembles the standard method when $g$ lower bounds $f$.) The gradient of $\mathbb{E}g$ is then *corrected* by a stochastic vector. The variance of the correction is smaller than that of the original stochastic approximation in Sec. 3, since the function $f(\theta)$ is close to $\hat{a}g(\theta)$. The gradient of $\mathbb{E}g$ can be thought of as an initial guess, followed by a stochastic correction which ensures that we are optimizing the variational objective function.

### 4.2. The stochastic search case

We have introduced a control variate for $f(\theta)$, but in fact we would like to minimize the variance of the vector $f(\theta)\nabla_\psi \ln q(\theta|\psi)$ in Eq. (6). In this case, the control variate becomes $g(\theta)\nabla_\psi \ln q(\theta|\psi)$ and we have the following modification.

Let $\psi_k$ be the $k$th dimension of $\psi$. Then for each dimension the discussion in Sec. 4.1 carries through, but for $f\frac{\partial \ln q}{\partial \psi_k}$ and $g\frac{\partial \ln q}{\partial \psi_k}$ instead of $f$ and $g$. The variance of each dimension again follows Eq. (9), and we seek an $a$ to minimize the sum of these equations. This results in the optimal value

$$a = \sum_k \text{Cov}(f\tfrac{\partial \ln q}{\partial \psi_k}, g\tfrac{\partial \ln q}{\partial \psi_k}) / \sum_k \text{Var}(g\tfrac{\partial \ln q}{\partial \psi_k}),$$

which we approximate using samples. We summarize stochastic search variational Bayes in Algorithm 1.

## 5. Stochastic search VB for two models

We next illustrate stochastic search variational inference on logistic regression and a finite approximation to the hierarchical Dirichlet process (Teh et al., 2007). For logistic regression, we will consider two control variates, one of which is a lower bound and the other of which is not a bound. For the finite HDP, we will consider a piecewise control variate, one part being an upper bound on the original function.

### 5.1. Logistic regression

Binary logistic regression takes in $d$-dimensional data vectors $x_n$ and predicts the class $y_n \in \{-1, 1\}$ to which each belongs. The parameter is $\theta \in \mathbb{R}^d$ and the prediction law is $\Pr(y_n|x_n, \theta) = \sigma(y_n x_n^T \theta)$ where $\sigma(\cdot)$ is the sigmoid function, $\sigma(b) = (1+e^{-b})^{-1}$. Bayesian logistic regression places a prior distribution on the coefficient vector, $\theta \sim \text{Normal}(0, cI)$. For inference we define a Gaussian variational $q$ distribution

$$q(\theta) = \text{Normal}(\mu, \Sigma). \quad (13)$$

The variational lower bound for this model is

$$\mathcal{L} = \sum_{n=1}^{N} \mathbb{E}_q[\ln \sigma(y_n x_n^T \theta)] + \mathbb{E}_q[\ln p(\theta) - \ln q(\theta)]. \quad (14)$$

The expectations of $f_n(y_n, x_n; \theta) := \ln \sigma(y_n x_n^T \theta)$ are intractable. One approach to avoiding this issue is to forgo variational inference and use Laplace's method to approximate $q$. This method sets $\mu$ to the MAP solution, and $\Sigma^{-1}$ to the negative Hessian of the log joint likelihood evaluated at $\mu$. Another is to lower bound $f_n$ with the bound in, e.g., Jaakkola & Jordan (2000), which allows for closed-form variational inference. We consider this bound as a control variate.

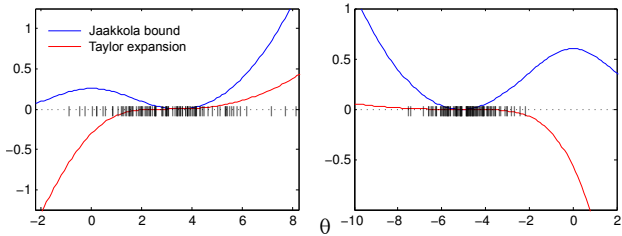

*Figure 1.* Approximation error between $\ln \sigma(\theta)$ and the two control variates considered. The mean and variance of $q$ used in these examples are (left) $\mu = 3$, $\sigma^2 = 3$ and (right) $\mu = -5$, $\sigma^2 = 1$. We show 100 samples from these $q$ distributions, at which points the functions would be evaluated for the stochastic gradient (for $a = 1$). The Taylor expansion is closer to the true function at the region of interest as defined by $q$. The benefit of this is that fewer samples will be necessary to approximate the gradient.

**A lower bound control variate.** The lower bound for $f_n$ developed by Jaakkola & Jordan (2000) is a useful control variate for variational logistic regression. For each pair $(x_n, y_n)$, this bound takes an auxiliary parameter $\xi_n > 0$ and has the form

$$g_n(y_n, x_n; \theta, \xi_n) = \ln \sigma(\xi_n) + \frac{1}{2}(y_n x_n^T \theta - \xi_n) \\ - \lambda(\xi_n)((x_n^T \theta)^2 - \xi_n^2). \quad (15)$$

We have $\lambda(\xi_n) = (2\sigma(\xi_n) - 1)/(4\xi_n)$. We select this bound for illustrative purposes, but any lower bound will work in principle. For a multivariate Gaussian $q$ distribution, having a quadratic term in $g$ is essential for stochastically learning a full covariance matrix. In general, tighter bounds will require fewer samples, but for some functions finding tight bounds may require much effort. We next consider a general purpose control variate that can help in this case.

**Control variate delta method.** We also consider the second-order Taylor expansion of $f$ as a control variate. The second-order Taylor expansion often accurately approximates a function of interest, and when used alone is known as the delta method. In addition to accuracy, the quadratic approximation of the delta method results in a function for which the expectation with respect to $q$ is very likely to be analytic.

The delta method arguably should not be used for mean-field variational inference because the second-order Taylor expansion is not a lower bound. On the other hand, the first-order Taylor expansion often is a lower bound. Therefore, though their bounds are typically loose, first-order approximations are commonly employed for MFVB. An advantage of the proposed stochastic search algorithm is that second-order methods can now be used as a control variate to $(i)$ more accurately approximate the function of interest, and $(ii)$ significantly reduce the variance of the stochastic gradient. We call this approach of using Taylor expansion control variates the *control variate delta method*.

We consider a second-order Taylor expansion at $\hat{\mu}$, the current mean of $q$, for approximating $\ln \sigma(y_n x_n^T \theta)$. Letting $\sigma_n := \sigma(y_n x_n^T \hat{\mu})$, this control variate is

$$g_n(y_n, x_n; \theta, \hat{\mu}) = \ln \sigma_n + y_n(1 - \sigma_n)(\theta - \hat{\mu})^T x_n \quad (16) \\ - \frac{1}{2}\sigma_n(1 - \sigma_n)(\theta - \hat{\mu})^T x_n x_n^T (\theta - \hat{\mu}).$$

As with the Jaakkola & Jordan (2000) bound, this control variate contains a quadratic term that helps in learning the covariance matrix of $q$.

We compare these control variates in Figure 1. In these plots we show the difference $f_n - g_n$ for two specific $q$ distributions, and with $x = 1$. We also show 100 samples from $q$, which indicates the regions where these functions would be evaluated (for $a = 1$). The plots show that the second-order Taylor expansion approximates $f_n$ significantly better where it matters; we support this conclusion with the experiments in Sec. 6.

### 5.2. Hierarchical Dirichlet processes

We also investigate a stochastic search VB algorithm for an approximation to the hierarchical Dirichlet process (Teh et al., 2007). We focus on the two-level generative structure using finite dimensional Dirichlet priors as an approximation to the infinite dimensional process—in the limit the HDP is recovered. In this finite process, a top-level Dirichlet-distributed probability vector $\theta$ parameterizes the Dirichlet distribution for $d = 1, \ldots, D$ second-level probability vectors,

$$(\pi_{d1}, \ldots, \pi_{dK}) \stackrel{iid}{\sim} \text{Dirichlet}(\beta\theta_1, \ldots, \beta\theta_K), \\ (\theta_1, \ldots, \theta_K) \sim \text{Dirichlet}(\tfrac{\alpha}{K}, \ldots, \tfrac{\alpha}{K}). \quad (17)$$

In topic models, these $\pi_d$ vectors are often used as distributions on word distributions. In this section, we focus solely on the generic hierarchical structure in Eq. (17). We define the approximate posterior of $\theta$ as

$$q(\theta) = \text{Dirichlet}(c_1, \ldots, c_K). \quad (18)$$

The part of the lower bound associated with $\theta$ is

$$\mathcal{L}_\theta = \sum_k \beta \mathbb{E}_q[\theta_k] \sum_d \mathbb{E}_q[\ln \pi_{dk}] - \sum_k D \mathbb{E}_q[\ln \Gamma(\beta\theta_k)] \\ + \mathbb{E}_q[\ln p(\theta) - \ln q(\theta)]. \quad (19)$$

The expectation $\mathbb{E}_q[\ln \Gamma(\beta\theta_k)]$ is intractable for each $k$. We use a stochastic approximation, and introduce two control variates for this function, depending on the current expected value of $\beta\theta_k$.

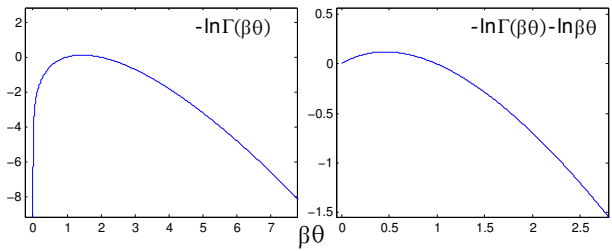

*Figure 2.* (left) The intractable function in the HDP. (right) the difference after introducing a control variate and setting $a = 1$. Since $-\ln\Gamma(\beta\theta) - \ln\beta\theta = -\ln\Gamma(\beta\theta+1)$, the right figure is the left figure shifted by one unit to the left and truncated at zero. The very large variance near zero (where most values of $\beta\theta$ will lie) has been significantly reduced. For larger values of $\beta\theta$, we use a first-order Taylor approximation at $\beta\mathbb{E}_q\theta$ of the nearly linear function.

As $\beta\theta_k$ approaches zero, the function $f_k(\beta\theta_k) = -\ln\Gamma(\beta\theta_k)$ diverges to $-\infty$. By construction of the Dirichlet prior, many values of $\theta_k$ will be very small. (In the infinite limit, there are an infinite number of such values smaller than any $\delta > 0$.) The variance in this region is massive—when computer precision becomes an issue it can be infinite (see Figure 2).

We propose the control variate $g_k(\beta\theta_k) = \ln\beta\theta_k$, with $\mathbb{E}_q[\ln\theta_k] = \phi(c_k) - \phi(\sum_j c_j)$ where $\phi(\cdot)$ is the digamma function. This control variate not only correlates well with $f_k$, but if $a = 1$, $\lim_{\beta\theta_k \to 0} f_k - ag_k = 0$, as shown in Figure 2. This results from the equality

$$-\ln\Gamma(\beta\theta_k) - \ln\beta\theta_k = -\ln\Gamma(\beta\theta_k + 1). \quad (20)$$

For all other values of $a$, this equality does not hold, and the difference $f_k - ag_k$ diverges as $\beta\theta_k \to 0$. For this model, we can thus give the optimal value of $a$ in advance, and we set $a = 1$.

From Figure 2, we also see that the approximation gets worse when $\beta\theta_k$ gets large, which can occur for a few highly probable dimensions when $\beta$ is large. Since $f_k$ is approximately linear in this regime, we use a first-order Taylor expansion of $f_k$ about the mean $\bar{\theta}_k = \mathbb{E}_q[\theta_k]$ as a control variate. This gives the following two control variates,

$$g_k = \ln\beta\theta_k, \quad 0 < \beta\bar{\theta}_k < \kappa_1, \quad (21)$$
$$g_k = -\ln\Gamma(\beta\bar{\theta}_k) - \beta(\theta_k - \bar{\theta}_k)\phi(\beta\bar{\theta}_k), \quad \beta\bar{\theta}_k > \kappa_2.$$

Since $f_k$ is concave, this second control variate is an *upper bound* on $\mathcal{L}_\theta$ without the stochastic correction. We discuss the boundaries $\kappa_1$ and $\kappa_2$ in Sec. 6.

Thus far, we've focused mainly on reducing the variance induced by $f_k$, but in Sec. 4.2 we noted that $\nabla \ln q$ introduces variance to the Monte Carlo integral as well. This suggests that we should look at other parts of the integral for potential variance reduction. We briefly show how this can be done for the HDP.

The lower bound in Eq. (19) contains a sum of $K$ intractable integrals over the probability simplex $\Delta_K$. We perform separate stochastic approximations of each gradient. Using the fact that each gamma function is over a single dimension of the simplex, for a function of $\theta_k$ the variables $\theta_{i\neq k}$ will integrate out. In this case, marginalizing a Dirichlet distribution to a single dimension yields a beta distribution. That is,

$$\int_{\theta \in \Delta_K} \ln\Gamma(\beta\theta_k)q(\theta|c)d\theta = \int_0^1 \ln\Gamma(\beta\theta_k)q'_k(\theta_k|c)d\theta_k,$$

where $q'_k(\theta_k|c) = \text{Beta}(\theta_k|c_k, \sum_{i\neq k} c_i)$.

We can choose which of these integrals to stochastically approximate for gradient ascent. However, the stochastic gradient using $q'_k$ results in significantly less variance than for $q$ since $\theta_k^{(s)}$ will be near zero; the vector $\nabla_c \ln q'_k$ has $K - 1$ entries containing $\ln(1 - \theta_k^{(s)}) - \mathbb{E}_q[\ln(1 - \theta_k)]$, while these values will be $\ln\theta_i^{(s)} - \mathbb{E}_q[\ln\theta_i]$ for $i = 1, \ldots, K$ when using $\nabla_c \ln q$.

## 6. Experiments

We perform experiments using stochastic search VB for binary classification with logistic regression and for topic modeling with the approximate HDP. We next give the details of the experiments we perform and the data sets and algorithms used for comparison.

**Data and set-up.** For logistic regression, we use five data sets from the UCI repository: Iris, Pima, SPECTF, Voting and WDBC. These data sets range from 150 to 768 labeled examples living in 5 to 45 dimensions, including a dimension of all ones to account for offset. We perform experiments with stochastic search variational inference using the two control variates discussed in Sec. 5.1. We compare with two additional methods for posterior approximation: variational inference with the Jaakkola & Jordan (2000) bound and Laplace's method. We evaluate performance on the true variational objective function in Eq. (14) using each posterior approximation.

For the HDP topic model, we use 8,000 documents with 3,012 vocabulary size from *The New York Times*. We compare with $(i)$ a point estimate of the top-level probability vector using a delta $q$ distribution, and $(ii)$ fixing the top-level distribution to the uniform vector, which is equivalent to LDA (Blei et al., 2003). We perform experiments for different corpus sizes, different values of $\beta$, and we set $K = 200$.

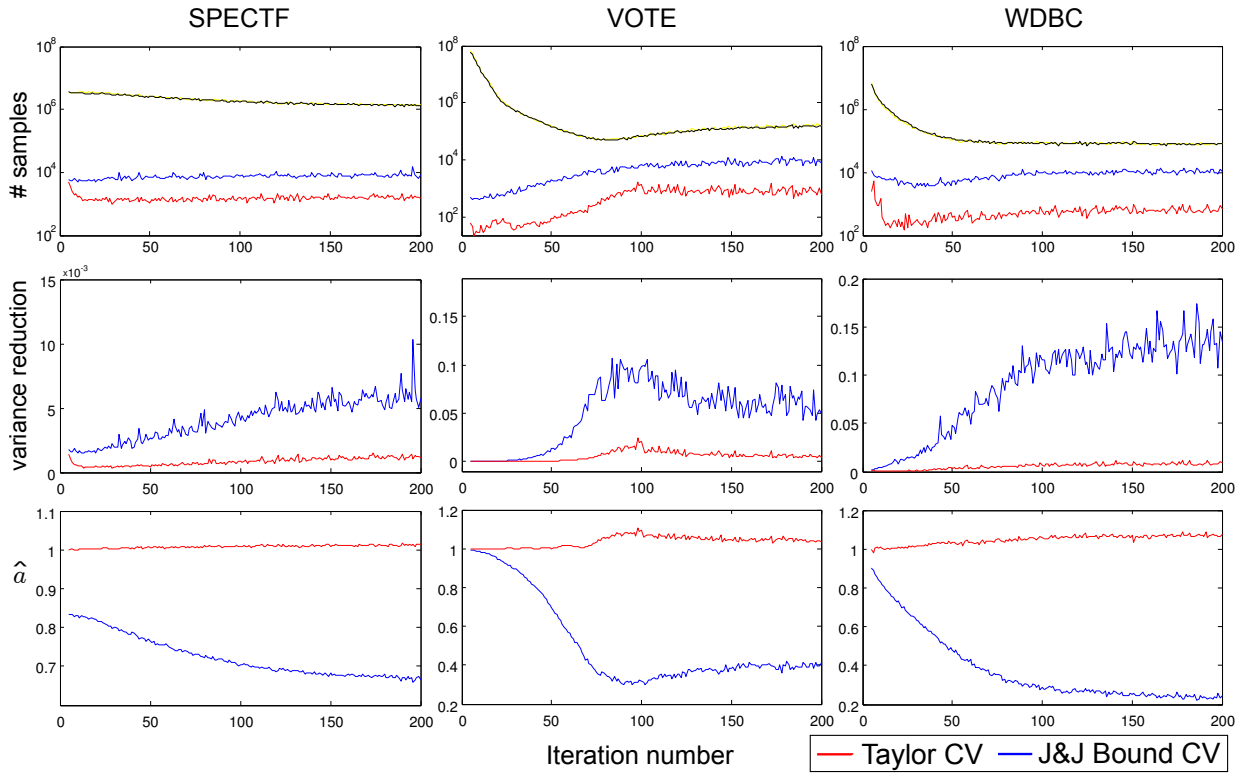

*Figure 3.* Experimental results for variational logistic regression. We compare the variance reduction obtained by the two control variates under consideration. (top row) The number of samples per iteration setting $\epsilon = 0.1$ in Algorithm 1. The yellow and black lines represent the estimated number that would be required without variance reduction according to each control variate. As expected, these curves overlap. (middle row) The variance reduction factor of Eq. (11). The selected control variates significantly reduce the variance. The second-order Taylor control variate is significantly better than the lower bound. (bottom row) The optimal scaling factor estimated from samples.

*Table 1.* Optimizing the variational objective function. The stochastic search methods (indicated by CV) significantly outperform the other methods toward this end. All values were calculated for the true lower bound in Eq. (14) using their respective posterior approximations.

| model|data | iris  | pima  | spectf | vote  | wdbc  |
|------------|-------|-------|--------|-------|-------|
| Taylor CV  | -7.9  | -3974 | -165   | -67.8 | -74.6 |
| J&J CV     | -7.9  | -3974 | -165   | -67.6 | -74.8 |
| Laplace    | -11.9 | -3985 | -170   | -70.5 | -80.0 |
| J&J bnd    | -11.5 | -3976 | -173   | -74.6 | -86.2 |

*Table 2.* Running time of each algorithm on each data set. We use the approximated number of samples required without a control variate to estimate the last value. The times are given in milliseconds (ms), seconds (s), minutes (m), hours (hr) and years (yr).

| model|data | iris  | pima  | spectf | vote  | wdbc  |
|------------|-------|-------|--------|-------|-------|
| Taylor CV  | 0.33s | 1.7m  | 20s    | 17s   | 11s   |
| J&J CV     | 0.42s | 18m   | 1.2m   | 1.2m  | 2.3m  |
| Laplace    | 21ms  | 29ms  | 94ms   | 20ms  | 0.10s |
| J&J bnd    | 64ms  | 88ms  | 0.13s  | 97ms  | 0.15s |
| SS no CV   | 2.4s  | >2yr  | 6.6hr  | 9hr   | 1.4hr |

**Logistic regression results.** In Table 1 we show the variational lower bound for each model on each data set. Since all algorithms return an approximation of the posterior distribution on the vector $\theta$, this comparison is meaningful and gives a measure of how close each posterior is to the true posterior. We see a considerable improvement for the stochastic algorithms (denoted by their control variate). Since both stochastic algorithms optimize the same objective, the performance should be the same.

We show performance details of the stochastic search VB algorithm in Figure 3 and Table 2. In Figure 3, we show the number of samples, the variance reduction factor and the scaling $\hat{a}$ as a function of iteration. We see that the control variates provide a major reduction in variance. Also, the Taylor expansion control variate (i.e., control variate delta method) requires significantly fewer samples than the bound control variate, which benefits the running time (see Table 2). While the non-sampling methods are faster, control variates make stochastic search VB a viable inference method when compared to the base algorithm of Sec. 3.

Table 3. The fraction of times that algorithm ⟨row⟩ was ranked ⟨column⟩ for the 32 different parameter/data size pairs using the variational lower bound.

| model \ rank | 1st | 2nd | 3rd |
|---|---|---|---|
| HDP-stochastic | 0.66 | 0.31 | 0.03 |
| HDP-point | 0.34 | 0.66 | 0 |
| LDA | 0 | 0.03 | 0.97 |

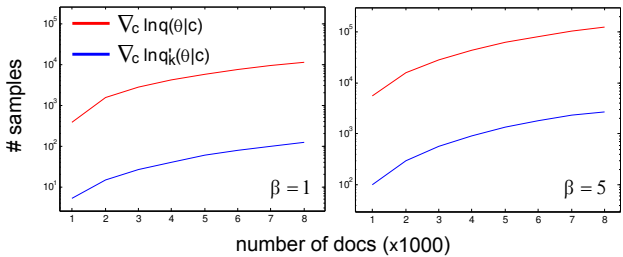

Figure 4. Average number of samples per iteration for the two equivalent gradient approximations, $\nabla_c \ln q$ vs $\nabla_c \ln q'_k$, where $q$ is the Dirichlet and $q'_k$ the beta distribution. Sampling is further reduced (see text for discussion).

**Hierarchical Dirichlet process results.** We fit topic models to *The New York Times* using different numbers of documents ($D = 1000$ to $8000$) and concentration parameter values $\beta \in \{1, 5, 10, 15\}$. As switch points for the two control variates, we set $\kappa_1 = 1$ and $\kappa_2 = 2$. We summarize our results in Table 3. In general, fitting a variational posterior on the top-level Dirichlet vector yielded a better posterior approximation than a point estimate and a $\theta$ fixed as uniform. However, this improvement was not as dramatic as for logistic regression.

In Figure 4, we show the number of samples required from the Dirichlet $q$ distribution to approximate the stochastic integral. We compare the two methods discussed in Sec. 5.2 for reducing the variance of the stochastic vector $\nabla_c \ln q$ by instead using $\nabla_c \ln q'_k$. We see a significant reduction in the number of samples. Experiments without control variates were not possible due to computer precision issues and the massive variance of $\ln \Gamma(\beta\theta)$ near zero.

## 7. Conclusion

We have presented stochastic search variational Bayes, a method for optimizing intractable variational objective functions such as those arising from non-conjugacy. The algorithm relies on a stochastic approximation of the gradient; we showed how control variates can significantly reduce the variance of this Monte Carlo integral. Since existing lower bounds can be recast as control variates, our approach is relevant to many existing MFVB algorithms. However, a lack of restrictions on control variates allows for other types of function approximations when a good bound is not readily available. We introduced the control variate delta method toward this end.

**Acknowledgements** J.P. and M.J. are supported by ONR grant number N00014-11-1-0688 under the MURI program. D.B. is supported by ONR N00014-11-1-0651, NSF CAREER 0745520, AFOSR FA9550-09-1-0668, the Alfred P. Sloan foundation, and a grant from Google.